\documentclass[runningheads]{llncs}
\usepackage{graphicx}
\usepackage{tikz}
\usepackage{comment} 
\usepackage{amsmath,amssymb} % define this before the line numbering.
\usepackage{color}

\usepackage{multirow}
\usepackage{bm}
\usepackage[tight,footnotesize]{subfigure}

\begin{document}
% \linenumbers
\pagestyle{headings}
\mainmatter
\def\ECCVSubNumber{3997}  % Insert your submission number here

\title{Self-Prediction for Joint Instance and Semantic Segmentation of Point Clouds} % Replace with your title

%\begin{comment}
%\titlerunning{ECCV-20 submission ID \ECCVSubNumber} 
%\authorrunning{ECCV-20 submission ID \ECCVSubNumber} 
%\author{Anonymous ECCV submission}
%\institute{Paper ID \ECCVSubNumber}
%\end{comment}
%******************

%\begin{comment}
\titlerunning{SP for Joint Instance and Semantic Segmentation}
\author{Jinxian Liu\inst{1,2}\thanks{Equal contribution. This work is done during their internships at Huawei Hisilicon.} \and
Minghui Yu\inst{1,2\star} \and
Bingbing Ni\inst{1,2}\protect\footnotemark[4] \and
Ye Chen\inst{1,2}}
\authorrunning{Liu et al.}
\institute{Shanghai Jiao Tong University\\
\email{\{liujinxian,1475265722,nibingbing,chenye123\}@sjtu.edu.cn}\\
              \and Huawei Hisilicon\\
\email{\{liujinxian1,yuminghui2,nibingbing,chenye17\}@hisilicon.com}
}
%\end{comment}
%******************
\maketitle

\renewcommand{\thefootnote}{\fnsymbol{footnote}} 
\footnotetext[4]{Corresponding author: Bingbing Ni.} 
\renewcommand{\thefootnote}{\arabic{footnote}}

\begin{abstract}
We develop a novel learning scheme named Self-Prediction for 3D instance and semantic segmentation of point clouds. Distinct from most existing methods that focus on designing convolutional operators, our method designs a new learning scheme to enhance point relation exploring for better segmentation. More specifically, we divide a point cloud sample into two subsets and construct a complete graph based on their representations. Then we use label propagation algorithm to predict labels of one subset when given labels of the other subset. By training with this Self-Prediction task, the backbone network is constrained to fully explore relational context/geometric/shape information and learn more discriminative features for segmentation. Moreover, a general associated framework equipped with our Self-Prediction scheme is designed for enhancing instance and semantic segmentation simultaneously, where instance and semantic representations are combined to perform Self-Prediction. Through this way, instance and semantic segmentation are collaborated and mutually reinforced. Significant performance improvements on instance and semantic segmentation compared with baseline are achieved on S3DIS and ShapeNet. Our method achieves state-of-the-art instance segmentation results on S3DIS and comparable semantic segmentation results compared with state-of-the-arts on S3DIS and ShapeNet when we only take PointNet++ as the backbone network.
\end{abstract}

\section{Introduction}
With the growing popularity of low-cost 3D sensors, e.g., LiDAR and RGB-D cameras, 3D scene understanding is tremendous demand recently due to its great application values in autonomous driving, robotics, augmented reality, etc. 3D data provides rich information about the environment, however, it is hard for traditional convolutional neural networks (CNNs) to process this irregular data. Fortunately, many ingenious works~\cite{Qi_2017_CVPR,DBLP:conf/nips/QiYSG17,dgcnn,DBLP:conf/cvpr/SuJSMK0K18,DBLP:conf/cvpr/XieLCT18,DBLP:conf/cvpr/ShenFYT18,DBLP:conf/cvpr/HuaTY18,DBLP:conf/nips/LiBSWDC18,DBLP:conf/cvpr/LiuFXP19,DBLP:conf/cvpr/LanYYD19,Zhang_2019_ICCV,Liu_2019_ICCV,Mao_2019_ICCV} are proposed to directly process point cloud, which is the simplest 3D data format. This motivates us to work with 3D point clouds.

The key to better understanding a 3D scene is to learn more discriminative point representations. To this end, many works~\cite{DBLP:conf/cvpr/LiuFXP19,dgcnn,DBLP:conf/eccv/WangSS18,DBLP:conf/eccv/XuFXZQ18,DBLP:conf/cvpr/LanYYD19} elaborately design various point convolution operators to capture semantic or geometric relation among points. DGCNN~\cite{dgcnn} proposes to construct a KNN graph and an operator named EgdeConv to process this graph, where semantic relation among points is explicitly modeled. RelationShape~\cite{DBLP:conf/cvpr/LiuFXP19} attempts to model geometric point relation in local areas, hence local shape information is captured. Other methods also share similar design philosophy. Although good segmentation performance is achieved by explicitly modeling points relation, lack of constraint/guidance on relation exploring limits the network from reaching its full potential. Hence a constraint is urgently needed to enforce/guide/encourage this relation exploring and helps the network learn more representative features.

3D Instance and semantic segmentation are two of the most important tasks in 3D scene understanding. Many works~\cite{DBLP:conf/3dim/TchapmiCAGS17,DBLP:conf/cvpr/GrahamEM18,Choy_2019_CVPR,DBLP:conf/cvpr/LandrieuS18,DBLP:conf/eccv/YeLHDZ18,DBLP:conf/cvpr/YiZWSG19,DBLP:conf/nips/YangWCHWMT19} tackle these two tasks separately. And some works~\cite{DBLP:conf/wacv/PhamHNY19,DBLP:conf/cvpr/QiLWSG18} address these two tasks in a serial fashion, where instance segmentation is usually formulated as a post-processing task of semantic segmentation. However, this formulation often gets a sub-optimal solution since the performance of instance segmentation highly depends on the performance of semantic segmentation. Actually, these two tasks could be associated and cooperate with each other as proved in ASIS~\cite{DBLP:conf/cvpr/WangLSSJ19} and JSIS3D~\cite{DBLP:conf/cvpr/PhamNHRY19}. They propose to couple these two tasks in a parallel fashion. ASIS makes instance segmentation benefit from semantic segmentation through learning semantic-aware instance embeddings. Semantic features of the points belonging to the same instance are fused to make more accurate semantic predictions. However, extra parameters and computation burden are introduced during inference. JSIS3D combines these two tasks in a simple way. They formulate it as a simple multi-task problem and just train the two tasks simultaneously. A multi-value conditional random fields model is proposed to jointly optimize class labels and object instances. However, it is a time consuming post-processing scheme and cannot be optimized end-to-end. Moreover, performance improvements achieved by ASIS and JSIS3D are both limited.

\begin{figure}[t]
\centering
\includegraphics[width=4.8in]{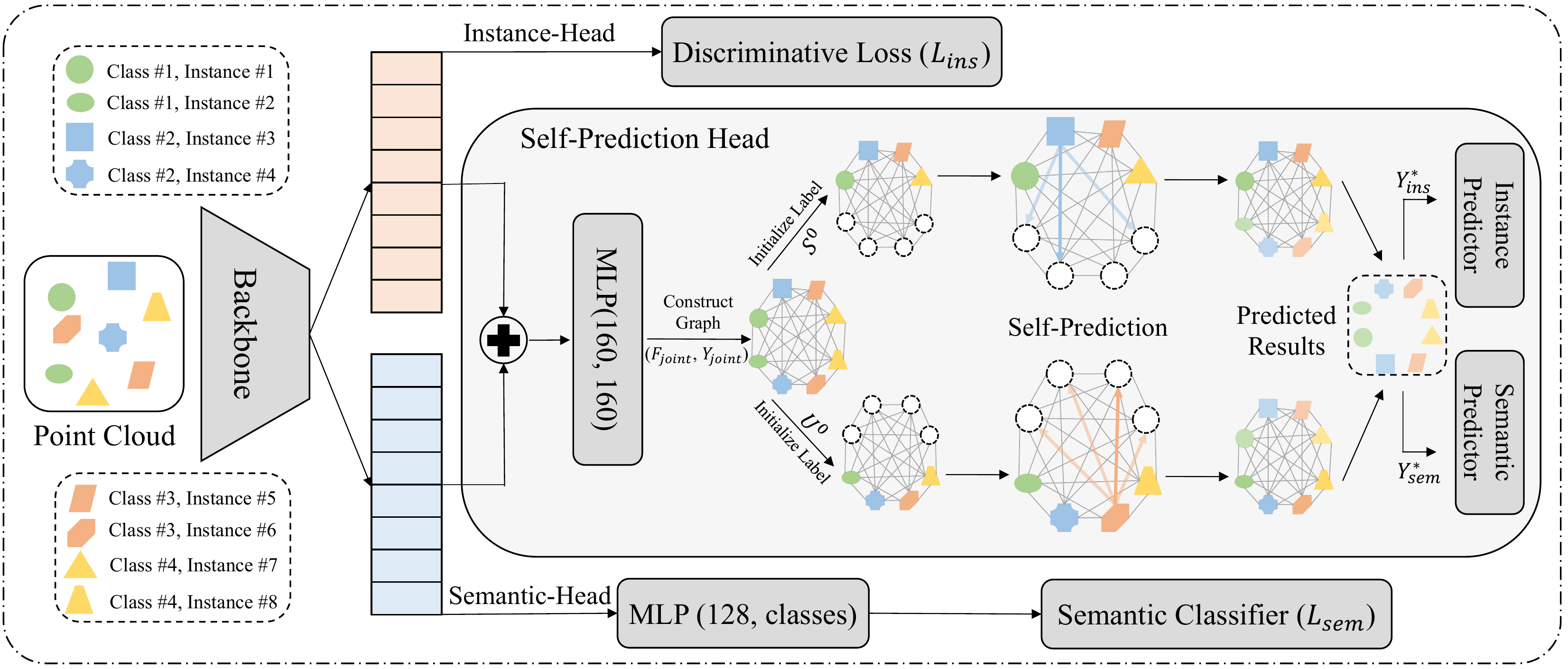}
\caption{The overall framework of our method. The input point cloud goes through a backbone network to extract instance and semantic features for instance and semantic segmentation respectively. These two features are then combined to construct a complete graph to perform bidirectional Self-Prediction in Self-Prediction head.}
\label{fig:framework}

\end{figure}

To address these two issues, we propose a novel learning scheme named Self-Prediction to constrain the network to fully capture point relation and a unified framework that equipped with this scheme to associate instance and semantic segmentation.
The framework of our method is shown in Figure~\ref{fig:framework}, which contains a backbone network and three heads named instance-head, semantic-head and Self-Prediction head respectively. The instance-head learns instance embeddings for instance clustering and the semantic-head outputs semantic embeddings for semantic prediction. 
In Self-Prediction head, the instance and semantic embeddings for each point are combined. We then concatenate semantic and instance labels to form a multi-label for every point. After that, we divide the point cloud into two groups with one group's labels being discarded. Given the combined embeddings of the whole point cloud and labels of one group, we construct a complete graph and then predict semantic and instance labels simultaneously for the other group using label propagation algorithm. It should be noted that bidirectional propagation among the two groups are performed. Through this procedure of multi-label Self-Prediction, the instance and semantic embeddings are associatively enhanced. The process of Self-Prediction incorporates embedding similarity of points, which enforces the network to explore effective relation among points and learn more discriminative representations. The three heads are jointly optimized at training time. During inference, our Self-Prediction head is discarded, and no computation burden and network parameters are introduced. Our framework is demonstrated to be general and effective on different backbone networks such as PointNet, PointNet++, etc. Significant performance improvements over baseline are achieved on both instance and semantic segmentation. By only taking PointNet++ as the backbone, our method achieves state-of-the-art instance segmentation results and comparable semantic segmentation results compared with state-of-the-art networks.

\section{Related Work}
\textbf{Instance Segmentation in 3D Point Clouds.}
A pioneer work for instance segmentation in 3D point clouds can be found in~\cite{DBLP:conf/cvpr/WangYHN18}, which uses similarity matrix to yield proposals followed by confidence map for pruning proposals and utilizes semantic map for assigning labels. ASIS~\cite{DBLP:conf/cvpr/WangLSSJ19} proposes to associate instance segmentation and semantic segmentation to achieve semantic awareness for instance segmentation. JSIS3D~\cite{DBLP:conf/cvpr/PhamNHRY19} introduces a multi-value CRF model to jointly optimize class labels and object instances. However, their performance is quite limited. Encouraged by the success of RPN and RoI, GSPN~\cite{DBLP:conf/cvpr/YiZWSG19} generates proposals by reconstructing shapes and proposes Region-based PointNet to get final segmentation results. 3D-SIS~\cite{DBLP:conf/cvpr/HouDN19} is also a proposal-based method. However, proposal-based methods are usually two-stages and need pruning proposals. 3D-BoNet~\cite{DBLP:conf/nips/YangWCHWMT19} directly predicts point-level masks for instances within detected object boundaries. It is single-stage, anchor free and computationally efficient. However, there is a limitation on adaptation to different types of input point clouds. In this work, we propose a unified framework equipped with an efficient learning scheme to simultaneously improve instance and semantic segmentation significantly.

\textbf{Semantic Segmentation in 3D Point Clouds.}
PointNet~\cite{Qi_2017_CVPR} is the first to directly consume raw point clouds which processes each point identically and independently and then aggregates them through global max pooling. It well respects order invariances of input data and achieves strong performance. Pointnet++~\cite{DBLP:conf/nips/QiYSG17} applies PointNet in a recursive way to learn local features with increasing contextual scales thus it achieves both robustness and detailed features.
Attention~\cite{DBLP:conf/cvpr/XieLCT18,DBLP:conf/cvpr/YangZNLLZT19,DBLP:conf/cvpr/WangHHZS19,DBLP:conf/cvpr/ZhaoJFJ19} has also been paied to aggregate local features effectively. RSNet~\cite{DBLP:conf/cvpr/HuangWN18} proposes a lightweight local dependency module to efficiently model local structures in point clouds, which is composed by slice pooling layers, RNN layers and slice unpooling layers. SPLATNET~\cite{DBLP:conf/cvpr/SuJSMK0K18} utilizes sparse bilateral convolutional layers to maintain efficiency and flexibility. PointCNN~\cite{DBLP:conf/nips/LiBSWDC18} explores $\mathcal{X}$-transformation to promote both weighting input point features and permutation of points into a latent and potentially canonical order.
Graph convolutions~\cite{DBLP:conf/cvpr/ShenFYT18,DBLP:conf/cvpr/LandrieuS18,DBLP:conf/eccv/WangSS18} are also proposed for improving semantic segmentation task. SPG~\cite{DBLP:conf/cvpr/LandrieuS18} adapts graph convolutional network on compact but rich representations of contextual relationship between object parts. SEGCloud~\cite{DBLP:conf/3dim/TchapmiCAGS17} combines advantages of neural network and conditional random field to get coarse to fine semantics on points. DGCNN~\cite{dgcnn} tries to capture local geometric structures by a new operation named EdgeConv, which generates edge features describing relation between a point and its neighbors. \cite{Jiang_2019_ICCV} shares the same idea of edge features, which constructs an edge branch to hierarchically integrate point features and edge features. Different from these methods, PointConv~\cite{DBLP:conf/cvpr/WuQL19} proposes a density re-weighted convolution which can closely approximate 3D continuous convolution on 3D point set. 
KPConv~\cite{Thomas_2019_ICCV} uses kernel points located in Euclidean Space to define the area where each kernel weight is applied, which well models local geometry. Our method can take most of these models as backbone network and achieve better segmentation performance.

\textbf{Label Propagation Alogrithms.}
Label propagation is derived from unsupervised learning. \cite{xiaojin2002learning} is an earlier attempt to address this issue where the labeled data act as sources that push out labels through unlabeled data and then developed by~\cite{zhou2004learning}, which introduces consistency assumption to guide label propagation. It is mentioned in~\cite{wang2007label} that since scale parameter $\sigma$ will affect the performance significantly, to address this issue, LNP uses overlapped linear neighborhood patches to approximate the whole graph. How to automatically learn optimal $\sigma$ is worthwhile exploring.~\cite{zhou2004learning} proposes to learn parameter $\sigma$ by minimum spanning tree heuristic and entropy minimization. Label propagation algorithms are designed to enhance models with unlabeled samples. Motivated by our intention of enhancing point relation exploring, we design a new learning scheme to restrict the results predicted by label propagation algorithm be identical with their ground truth. 

\section{Methodology}
We design a novel learning scheme named Self-Prediction to strengthen our backbone networks and learn more discriminative representations for better segmentation. Among a point cloud sample, this proposed scheme encourages the network to capture more effective relation between points by predicting the label of a part of points when given labels of rest points and all points embeddings. Equipped with Self-Prediction, a unified framework is proposed to combine instance and semantic segmentation and conduct these two tasks in a mutually reinforced fashion. The overall framework of our method is shown in Figure~\ref{fig:framework}. In this section, we first introduce our proposed general Self-Prediction scheme. Then we present how to use this scheme to conduct instance and semantic segmentation jointly, and describe the overall framework. Finally, we summarize the global optimization objectives of our method.

\subsection{Self-Prediction}
\label{sec:sp}
Self-Prediction is an auxiliary task paralleled with instance and segmentation tasks, and is designed to enforce backbone networks to learn more strong and discriminative representations. To get better segmentation performance, many existing works~\cite{dgcnn,DBLP:conf/cvpr/ZhaoJFJ19,DBLP:conf/eccv/WangSS18,Thomas_2019_ICCV} elaborately design convolution operators to capture relation, geometric and shape information contained in point clouds. Our common goal is to learn more discriminative representations. However, we take a new perspective. We think that if the learned representations can be utilized to predict instance/semantic labels of a part of a point cloud when given labels of rest points in a point cloud, it can be considered to have fully exploited the relation information and be representative enough. Hence we naturally formulate a Self-Prediction task, i.e., equally divide a point cloud into two groups, and then perform bidirectional prediction between the two groups given their representations. By constraining the network to perform well on Self-Prediction task, we get more strong features and perform better on specific tasks, i.e., instance and semantic segmentation.

Given a point cloud example that contains $N$ points $X = \{\mathbf{x_1}, \mathbf{x_2}, ..., \mathbf{x_N}\}$, each point $\mathbf{x_i} \in \mathbb{R}^{h}$ can be represented by coordinates, color, normal, etc. $h$ is the dimension of features of input point. For each point $x_i$, its class label is represented by a one-hot vector. We formulate a label matrix $\mathbf{Y}\in \mathcal{Y}$, where each row of matrix $\mathbf{Y}$ denotes the one-hot label of point $\mathbf{x_i}$ and $\mathcal{Y}$ denotes the set of $N \times C$ matrix ($C$ is the number of classes) with non-negative elements.

We equally divide a point cloud into two groups, i.e., $X_S =\{\mathbf{x_1}, \mathbf{x_2}, ..., \mathbf{x_M}\}$ with its label matrix $\mathbf{Y}_{1:M}$ and $X_U =\{\mathbf{x_{M+1}}, \mathbf{x_{M+2}}, ..., \mathbf{x_N}\}$ with its label matrix $\mathbf{Y}_{M+1:N}$. We use label propagation algorithm to perform bidirectional Self-Prediction between point subsets $X_S$ and $X_U$, i.e., propagating labels from $X_S$ to $X_U$ and from $X_U$ to $X_S$ inversely. Firstly, we construct a complete graph $\mathbf{W} \in  \mathbb{R}^{N \times N}$, each element of which is defined by Gaussian similarity function:

\begin{equation}
\mathbf{W}_{ij} = exp(-\frac{d(\varphi(\mathbf{x}_{i}),\varphi(\mathbf{x}_{j}))}{2 \sigma ^{2}}).
\end{equation}
$\varphi$ is the backbone network and $\varphi(\mathbf{x}_{i})$ denotes extracted features of point $x_i$. $d$ is Euclidean distance measure function and $\sigma$ is the length scale parameter used to adjust the strength neighbors. We set $\sigma$ to $1$ in all our experiments. Then we normalize the constructed graph by computing Laplacian matrix:
\begin{equation}
\mathcal{\mathbf{L}}=\mathbf{D}^{-1/2}\mathbf{W}\mathbf{D}^{-1/2},
\end{equation}
where $\mathbf{D}$ is a diagonal matrix with $\mathbf{D}_{ii}$ to be the sum of the $i$-th row of $\mathbf{W}$, i.e., $\mathbf{D}_{ii} = \sum_{j=1}^{N}\mathbf{W}_{ij}$. To predict labels of $X_U$ when given labels of $X_S$ and labels of $X_S$ when given labels of $X_U$ respectively, we have to prepare two initial label matrices $\mathbf{S^{0}}$ and $\mathbf{U^{0}}$ by padding $\mathbf{Y}_{1:M}$ and $\mathbf{Y}_{M+1:N}$ with zero vectors correspondingly. Specifically, $\mathbf{S^{0}}$ and $\mathbf{U^{0}}$ are represented by:
\begin{equation} \label{eqn2}
  \begin{split}
 &\mathbf{S^{0}}=[\mathbf{Y}_{1}^T,...,\mathbf{Y}_{M}^T, \mathbf{0}^T,...,\mathbf{0}^T]^T,\\ 
 &\mathbf{U^{0}}=[\mathbf{0}^T,...,\mathbf{0}^T,\mathbf{Y}_{M+1}^T,...,\mathbf{Y}_{N}^T]^T,
  \end{split}
\end{equation}
where $\mathbf{Y}_{i}$ denotes the $i$-th row of label matrix $\mathbf{Y}$. The Self-Prediction procedure is conducted by label propagation algorithm, the iterative version of which is as follows:
\begin{equation} \label{eqn2}
  \begin{split}
  &\mathbf{S^{(t+1)}} = \alpha \mathbf{LS^{(t)}} + (1-\alpha)\mathbf{S^{0}},\\
  &\mathbf{U^{(t+1)}} = \alpha \mathbf{LU^{(t)}} + (1-\alpha)\mathbf{U^{0}},
  \end{split}
\end{equation}
where $ \alpha$ is a parameter used to control the propagation proportion, i.e., how much the initial label matrix has effect on propagated results. Following the common setting~\cite{DBLP:conf/iclr/LiuLPKYHY19}, we set $ \alpha$ to $0.99$ in all our experiments. $\mathbf{S^{(t)}} \in \mathcal{Y}$ and $ \mathbf{U^{(t)}} \in \mathcal{Y}$ are the $t$-th iteration results. We will get the final results $\mathbf{S^*}$ and $\mathbf{U^*}$ by iterating Equation~\ref{eqn2} until convergence. In practice, we directly use the closed form of the above iteration version that proposed in~\cite{zhou2004learning} to get propagated/predicted results. We present the closed form expression as follows:
\begin{equation} \label{closedform}
  \begin{split}
  &\mathbf{S^{*}} = (\mathbf{I}-\alpha\mathbf{L})^{-1}\mathbf{S^{0}},\\
  &\mathbf{U^{*}} = (\mathbf{I}-\alpha\mathbf{L})^{-1}\mathbf{U^{0}},
  \end{split}
\end{equation}
 where $\mathbf{I}\in\mathbb{R}^{N\times N}$ is the identity matrix. It should be noted that $\mathbf{S^{*}}_{{M+1}:N}$ and  $\mathbf{U^{*}}_{{1}:M}$ are valid propagated results. We can predict label of $x_i$ by $\arg\max \mathbf{U^{*}}_{i}$ when $1<i\le M$ and $\arg\max \mathbf{S^{*}}_{i}$ when $M<i\le N$. We formulate the final self-predicted results $\mathbf{Y^{*}} \in \mathcal{Y}$ as:
 \begin{equation}
\mathbf{Y^{*}} = [\mathbf{U^{*}}_{{1}:M}^T, \mathbf{S^{*}}_{{M+1}:N}^T]^T.
\end{equation}
Finally, we use ground truth label matrix $\mathbf{Y}$ as supervised signal to train this Self-Prediction task.

\subsection{Associated Learning Framework}
As shown in Figure~\ref{fig:framework}, our proposed framework contains one backbone network and three heads. The backbone network can be almost all existing point cloud learning architectures. We take PointNet, PointNet++ as examples in our work. Based on the backbone network, three heads are utilized to perform instance segmentation, semantic segmentation and Self-Prediction task respectively.

Taken a point cloud $X$ as input, the backbone network output a feature matrix $F \in \mathbb{R}^{N \times H}$, where $H$ denotes dimension of output features. Instance-head takes $F$ as input and transform it into point-wise instance embeddings $F_{ins} \in \mathbb{R}^{N \times H_{ins}}$, where $H_{ins}$ is dimension of instance embeddings and set to 32 in all our experiments. We adopt the same discriminative loss function as~\cite{DBLP:conf/cvpr/WangLSSJ19} and~\cite{DBLP:conf/cvpr/PhamNHRY19} to supervise instance segmentation. If a point cloud example contains $K$ instances and the $k$-th ($k \in {1, 2, ... K}$) instance contains $N_k$ points, we denote $\mathbf{e}_j  \in \mathbb{R}^{H_{ins}}$ as the instance embedding of the $j$-th point and $\bm{\mu}_k  \in \mathbb{R}^{H_{ins}}$ as the mean embedding of the $k$-th instance. Hence the instance loss is written as :
\begin{equation}
	\mathcal{L}_{var} = \frac{1}{K} \sum_{k=1}^{K} \frac{1}{N_k} \sum_{j=1}^{N_k} \left[ {\lVert \bm{\mu}_k - \mathbf{e}_j \rVert}_2 - \delta_{\textrm{v}} \right]_{+}^2,
\label{eq:l_var}
\end{equation}
\begin{equation}
\mathcal{L}_{dist} = \frac{1}{K (K-1)} \mathop{\sum_{k = 1}^{K} \sum_{m = 1,{m \neq k}}^{K}} \left[ 2 \delta_{\textrm{d}} - {\lVert \bm{\mu}_{k} - \bm{\mu}_{m} \rVert}_2 \right]_{+}^2,
\end{equation}
\begin{equation}
\mathcal{L}_{reg} = \frac{1}{K} \sum_{k=1}^{K} {\lVert \bm{\mu}_{k} \rVert}_2,
\end{equation}

  \begin{equation}
  \label{l_ins}
\mathcal{L}_{ins} = \mathcal{L}_{var} + \mathcal{L}_{dist} + 0.001 \cdot \mathcal{L}_{reg}
\end{equation}
where $[x]_+ = max(0, x)$, $\delta_{\textrm{v}}$ and $\delta_{\textrm{d}}$ are margins for $\mathcal{L}_{var}$ and $\mathcal{L}_{dist}$ respectively. Instance labels are obtained by conducting mean-shift clustering~\cite{DBLP:journals/pami/ComaniciuM02} on instance embeddings during inference.

The semantic-head takes feature matrix $F$ as input and learns a semantic embedding matrix $F_{sem} \in \mathbb{R}^{N \times H_{sem}}$ to further perform point-wise classification that supervised by cross-entropy loss. $H_{sem}$ is dimension of point semantic embedding and set to 128 in all our experiments.

In Self-Prediction head, we combine instance and semantic embeddings and jointly self-predict instance and semantic labels. Specifically, we concatenate $F_{ins}$ and $F_{sem}$ along the axis of features and transform it into a joint embedding matrix $F_{joint} \in \mathbb{R}^{H_{joint}}$, where $H_{joint}$ is dimension of joint embeddings and set to 160 in all our experiments. For each point in $X$, we transform its semantic and instance label into one-hot form respectively. Instance label of each point denotes which instance it belongs to. This instance label is semantic-agnostic, i.e., we cannot infer the semantic label of a point from its instance label. We assume that a dataset contains $C_{sem}$ semantic classes and the input point cloud sample $X$ contains $C_{ins}$ instances. Then we denote the semantic label matrix and instance label matrix as $\mathbf{Y}_{sem} \in \mathcal{Y}_{sem}$ and $\mathbf{Y}_{ins} \in \mathcal{Y}_{ins}$ respectively, where $\mathcal{Y}_{sem}$ is the set of $N \times C_{sem}$ matrix with non-negative elements and $\mathcal{Y}_{ins}$ is the set of $N \times C_{ins}$ matrix with non-negative elements. Given the two label matrices, we formulate a multi-label matrix $\mathbf{Y}_{joint} \in \mathcal{Y}_{joint}$ by concatenating semantic label and instance label of each point, where $\mathcal{Y}_{joint}$ is the set of $N \times (C_{sem}+C_{ins})$ matrix with non-negative elements. In other words, one can infer which semantic class and instance each point belongs to from the $\mathcal{Y}_{joint}$. We finally carry out Self-Prediction described in Section~\ref{sec:sp} based on the joint feature matrix $F_{joint}$ and multi-label matrix $\mathbf{Y}_{joint}$. We slice the self-predicted results $\mathbf{Y}_{joint}^* \in \mathcal{Y}_{joint}$ into semantic results $\mathbf{Y}_{sem}^* \in \mathcal{Y}_{sem}$ and instance results $\mathbf{Y}_{ins}^* \in \mathcal{Y}_{ins}$, which are then supervised by semantic ground truth $\mathbf{Y}_{sem}$ and instance ground truth $\mathbf{Y}_{ins}$ respectively. It should be noted that our Self-Prediction is conducted among one point cloud sample every time, hence it does not matter that the meaning of instance label varies from sample to sample.

Instance-head, semantic-head and Self-Prediction head are jointly optimized. Instance-head and semantic-head are aimed to get segmentation results. Our proposed Self-Prediction head incorporates similarity relation among points and enforces the backbone to learn more discriminative representations. These three heads cooperate with each other and get better segmentation performance. We want to emphasize that our Self-Prediction head is discarded and only instance-head and semantic-head are used during inference, hence no extra computational burden and space usage are introduced. 

\subsection{Optimization Objectives}
\label{loss}
We train the instance-head with the instance loss $\mathcal{L}_{ins}$ that formulated in Equation~\ref{l_ins}. The semantic-head is trained by classical cross-entropy loss and supervised by semantic label $\mathbf{Y}_{sem}$, which is written as:
\begin{equation}
    \mathcal{L}_{sem} = - \frac{1}{N} \sum_{i=1}^N [\mathbf{Y}_{sem}]_i \log \mathbf{p}_i,
\end{equation}
where $\mathbf{p}_i$ denotes output probability distribution computed by softmax function. 

Given the jointly self-predicted results $\mathbf{Y}_{ins}^*$ and $\mathbf{Y}_{sem}^*$, we train our Self-Prediction head also by cross-entropy loss, which is formulated as:
\begin{equation}
    \mathcal{L}_{sp} = - \frac{1}{N} \sum_{i=1}^N([\mathbf{Y}_{ins}]_i * \log \mathbf{q}_i + [\mathbf{Y}_{sem}]_i * \log \mathbf{r}_i),
\end{equation}
where $\mathbf{q}_i $ and $\mathbf{r}_i $ are output probability distribution (computed by softmax) of the $i$-th row of $\mathbf{Y}_{ins}^*$ and $\mathbf{Y}_{sem}^*$ respectively. The output probability distribution is also computed by softmax function.

The three head are jointly optimized and the overall optimization objective is a weighted sum of above three losses:
\begin{equation} 
    \mathcal{L} = \mathcal{L}_{ins} + \mathcal{L}_{sem} + \beta \mathcal{L}_{sp},
\end{equation}
where $\beta$ is used to balance contributions of the three above terms such that they contribute equally to the overall loss. $\beta$ is set to $0.8$ in all our experiments.

\section{Experiments}
\subsection{Experiment Settings}
\noindent{\textbf{Datasets}}\quad Stanford 3D Indoor Semantics Dataset (S3DIS) is a large scale real scene segmentation benchmark and contains 6 areas with a total of 272 rooms. Each 3D RGB point is annotated with an instance label and a semantic label from 13 categories. Each room is typically parsed to about 10-80 object instances. ShapeNet part dataset contains 16681 samples from 16 categories. There are totally 50 parts, and each category contains 2-6 parts. The instance annotations are got from~\cite{DBLP:conf/cvpr/WangYHN18}, which is used as ground truth instance label. 

\noindent{\textbf{Evaluation Metrics}}\quad On S3DIS dataset, following the common evaluation settings, we validate our method in a 6-fold cross validation fashion over the 6 areas, i.e., 5 areas are used for training and the left 1 area for validation each time. Moreover, test results on Area 5 are reported individually due to no overlaps between Area 5 and left areas, which is a better way to show generalization ability of methods. For evaluation of semantic segmentation, we use mean IoU (mIoU) across all the categories, class-wise mean of accuracy (mAcc) and point-wise overall accuracy (oAcc) as metrics. We take the same evaluation metric as~\cite{DBLP:conf/cvpr/WangLSSJ19} for instance segmentation. Apart from common used metric mean precision (mPrec) and mean recall (mRec) with IoU threshold 0.5, coverage and weighted coverage (Cov, WCov)~\cite{DBLP:conf/iccv/LiuJFU17,DBLP:conf/cvpr/RenZ17,DBLP:conf/cvpr/ZhuoSHL17} are taken. Cov is the average instance-wise IoU between prediction and ground truth. WCov means Cov that is weighted by the size of the ground truth instances. On ShapeNet, part-averaged IoU (pIoU) and mean per-class pIoU (mpIoU) are taken as evaluation metrics for semantic segmentation. Following~\cite{DBLP:conf/cvpr/WangYHN18,DBLP:conf/cvpr/WangLSSJ19}, we only provide qualitative results of part instance segmentation on ShapeNet.

\noindent{\textbf{Implementation Details}}\quad For experiments on S3DIS, we follow the same setting as PointNet~\cite{Qi_2017_CVPR}, where each room is split into blocks of area $1m\times 1m$. Each 3D point is represented by a 9-dim vector, (XYZ, RGB and normalized locations as to the room). We sample 4096 points for each block during training and all points are used for testing. We have mentioned above that we construct a graph and then divide the point cloud into two groups to perform Self-Prediction in Self-Prediction head. In practice, we partition the point cloud into more than two groups for acceleration. Specifically, we divide every block equally into 8 groups according to their instance labels, i.e., guarantee points of each instance are averagely distributed in every group. As a result, points of each semantics are also averagely distributed in every group. And then 4 pairs are randomly paired to conduct Self-Prediction. %Note that we only divide the point cloud in Self-Prediction head.
We train all models on S3DIS for 100 epochs with SGD optimizer and batch size 8. The base learning rate is set to 0.01 and divided by 2 every 20 epochs. For instance head, we set $\delta_{\textrm{v}}$ to $0.5$ and $\delta_{\textrm{d}}$ to $1.5$ following the same setting as ~\cite{DBLP:conf/cvpr/WangLSSJ19} and~\cite{DBLP:conf/cvpr/PhamNHRY19}. The loss weight coefficient $\beta$ for $\mathcal{L}_{sp}$ is set to $0.8$. BlockMerging algorithm~\cite{DBLP:conf/cvpr/WangLSSJ19,DBLP:conf/cvpr/PhamNHRY19} is used to merge instances from different blocks during inference, and bandwidth is set to $0.8$ for mean-shift clustering. 

For experiments on ShapeNet, input point cloud is represented only by coordinates. In Self-Prediction head, input point cloud is divided into 4 groups. We train all models for 200 epochs with Adam optimizer and batch size 16. The base learning rate is set to 0.001 and divided by 2 every 20 epochs. Other settings are the same as experiments conducted on S3DIS.

\subsection{Segmentation Results on S3DIS}
\label{sec:results on s3dis}
We report instance and semantic segmentation results in Table~\ref{tab:ins_bs} and Table~\ref{tab:sem_bs} respectively, where results of Area 5 and 6-fold cross validation are all shown. Baseline results in tables denote that we train our backbone network with only instance-head and semantic-head. All baseline results for PointNet and PointNet++ in the table are got from vanilla results of~\cite{DBLP:conf/cvpr/WangLSSJ19}, which are almost the same as ours. In all tables, InsSem-SP denotes complete version of our method, i.e., performing instance and semantic Self-Prediction jointly. To prove effectiveness of our proposed Self-Prediction scheme and our associated framework more clearly, we report the results of Ins-SP and Sem-SP in Table~\ref{tab:ins_bs} and Table~\ref{tab:sem_bs} respectively. Ins-SP means that we only perform instance Self-Prediction by taking $\mathbf{F}_{ins}$ and $\mathbf{Y}_{ins}$ as input. Sem-SP means that we only perform semantic Self-Prediction by taking $\mathbf{F}_{sem}$ and $\mathbf{Y}_{sem}$ as input. 
\vspace{-0.5cm}
\begin{table}[!ht]
\begin{center}
\small 
\setlength{\tabcolsep}{3.8pt}
\begin{tabular}{cccccc}
\hline
\hline
 Backbone & Method    &  mPrec & mRec  & mCov    & mWCov   \\
\hline
\hline
\multicolumn{6}{c}{Rseults on Area 5} \\
\hline
\multirow{4}{*}{PN} 
  & Baseline~\cite{DBLP:conf/cvpr/WangLSSJ19}  & 42.3 & 34.9 & 38.0 & 40.6\\
  & ASIS~\cite{DBLP:conf/cvpr/WangLSSJ19}  &  44.5 & 37.4 &40.4 &43.3 \\
  & Ours (Ins-SP) &  48.2 & 39.9 &44.7 &47.6\\
  & Ours (InsSem-SP) & {\bf 51.1} & {\bf 43.6}& {\bf 49.2}  & {\bf51.8} \\
 \hline
  \multirow{4}{*}{PN++}  
  & Baseline~\cite{DBLP:conf/cvpr/WangLSSJ19} & 53.4 & 40.6  & 42.6 & 45.7 \\
  & ASIS~\cite{DBLP:conf/cvpr/WangLSSJ19} &  55.3 & 42.4& 44.6  & 47.8  \\
  & Ours (Ins-SP)  &58.9&46.3&52.8&54.9\\
  & Ours (InsSem-SP)& {\bf 60.1} & {\bf 47.2}& {\bf 54.1}  & {\bf 56.3} \\
  
\hline 
\hline
\multicolumn{6}{c}{Results 6-fold CV} \\
\hline
 \multirow{4}{*}{PN} 
 & Baseline~\cite{DBLP:conf/cvpr/WangLSSJ19}& 50.6 & 39.2 & 43.0 & 46.3  \\
 &  ASIS~\cite{DBLP:conf/cvpr/WangLSSJ19} & { 53.2} & { 40.7}  & { 44.7}  & { 48.2} \\
   & Ours (Ins-SP) &55.1&44.3&48.9&50.1\\
  & Ours (InsSem-SP)&{\bf 56.6} & {\bf 45.9}& {\bf 51.8}  & {\bf 52.2}\\
 \hline
  \multirow{4}{*}{PN++}  
  & Baseline~\cite{DBLP:conf/cvpr/WangLSSJ19} & 62.7 & 45.8  & 49.6 & 53.4\\
 & ASIS~\cite{DBLP:conf/cvpr/WangLSSJ19}& { 63.6} & { 47.5}  & { 51.2}  & { 55.1} \\
   & Ours (Ins-SP) &65.9&53.2&58.0&60.7\\
  & Ours (InsSem-SP)& {\bf 67.5} & {\bf54.6}  & {\bf 60.4}  & {\bf 63.0} \\
\hline
\end{tabular}
\end{center}
\caption{Instance segmentation results on S3DIS dataset.}
\label{tab:ins_bs}
\end{table}
\vspace{-0.6cm}

\begin{figure}[h!]
        \begin{center}
        \subfigure[Instance segmentation]{\includegraphics[width=0.49\textwidth]{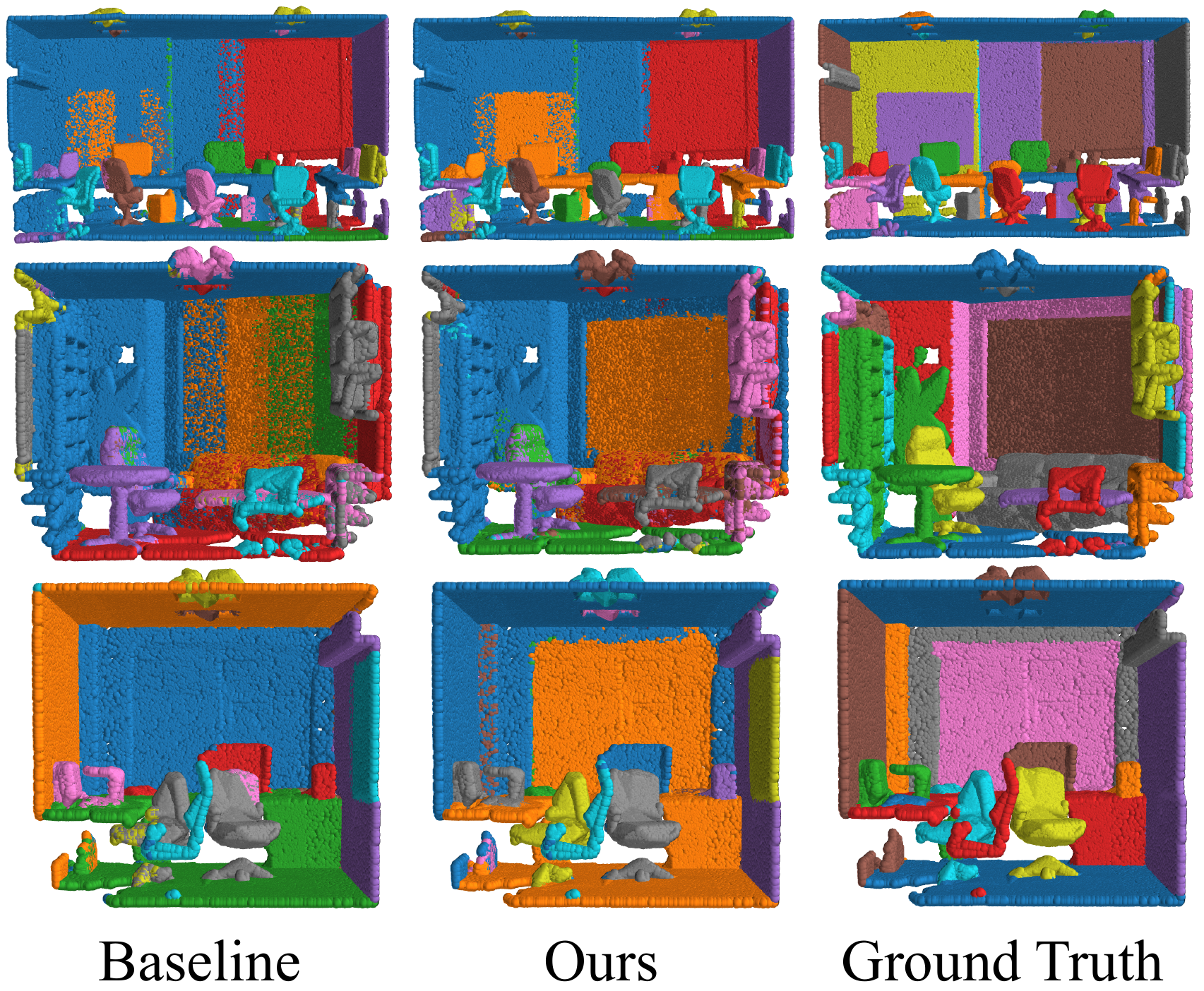}}
        \subfigure[Semantic segmentation]{\includegraphics[width=0.49\textwidth]{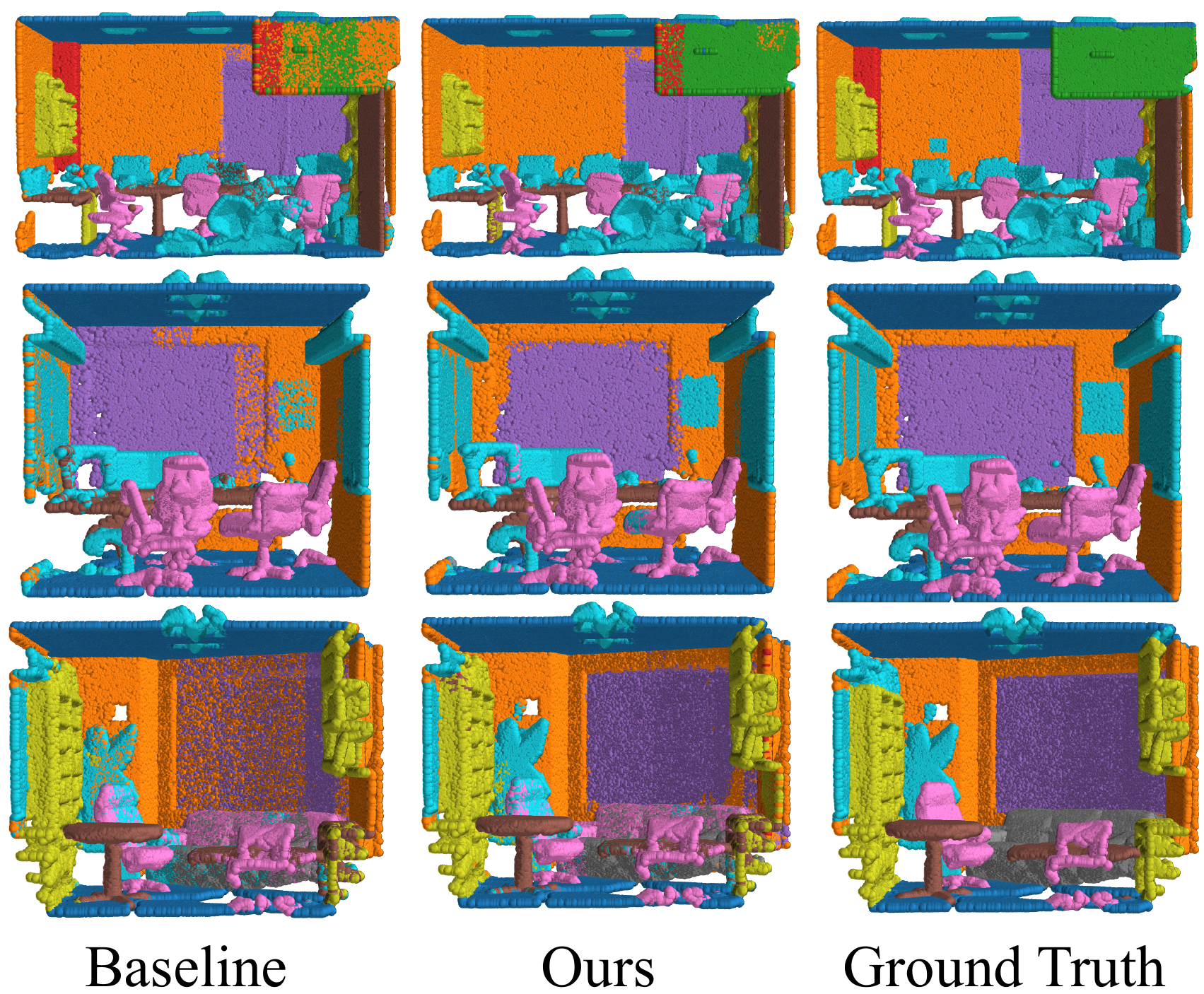}}
       \vspace{-0.2cm}
        \caption{Visualization results of instance and semantic segmentation. Our method obviously performs better than baseline. Best viewed in color.}
        \label{fig:shrec-trans}
            \label{vis_s3dis}
        \end{center}
\end{figure}

From Table~\ref{tab:ins_bs} and Table~\ref{tab:sem_bs}, we can observe that our method improves the baseline based on all three backbone networks on both instance and semantic segmentation tasks significantly. For example, our method improve baseline by 8.3 mPrec, 8.7 mRec, 11.2 mCov, 11.2 mWCov in instance segmentation and 7.7 mIoU, 9.5 mAcc, 3.9 oAcc in semantic segmentation on Area 5 when we use PointNet as backbone. Effectiveness of proposed Self-Prediction scheme is fully proved by comparing the results of Ins-SP with baseline in Table~\ref{tab:ins_bs} and the results of Sem-SP with baseline in Table~\ref{tab:sem_bs}. Moreover, performance is further improved when we conduct instance and semantic Self-Prediction jointly. In Figure~\ref{vis_s3dis}, we show some visualization results of baseline and Ours (InsSem-SP) based on PointNet++. We observe that our method achieves obvious more accurate predictions and performs better instance/semantic class boundaries.

\vspace{-0.5cm}
\begin{table}[!hbt]
\small 
\begin{center}
\begin{tabular}{cc|ccc|ccc}
\hline
\hline
    Backbone & Method    & mIoU  & mAcc    & oAcc& mIoU  & mAcc    & oAcc   \\
\hline
\hline 
&& &Area 5&&&6-fold CV& \\
\hline
\multirow{4}{*}{PN}  
 & Baseline~\cite{DBLP:conf/cvpr/WangLSSJ19}& 44.7 & 52.9  & 83.7 & 49.5 & 60.7  & 80.4 \\
 & ASIS~\cite{DBLP:conf/cvpr/WangLSSJ19} &46.4 &  55.7  & 84.5 & { 51.1} &  { 62.3}  & { 81.7}\\
 & Ours (Sem-SP) & 48.0 &  58.6  &  85.5 &52.3&64.5&83.0\\
 & Ours (InsSem-SP)&{\bf 52.4} &{\bf 62.4}&{\bf 87.6}  &{\bf 54.8} &{\bf 67.4} &{\bf 84.8}\\ 
\hline 
\multirow{4}{*}{PN++} 
 & Baseline~\cite{DBLP:conf/cvpr/WangLSSJ19} & 50.8 & 58.3  & 86.7 & 58.2 &69.0  & 85.9\\
 & ASIS~\cite{DBLP:conf/cvpr/WangLSSJ19}  &  53.4 & 60.9  & 86.9 &59.3&70.1&86.2\\
  & Ours (Sem-SP) &55.9&63.6&87.3 &61.1&72.2&87.3\\
  & Ours (InsSem-SP)&{\bf 58.8} &{\bf 65.9}&{\bf 89.2}&{\bf64.1} &{\bf74.3} &{\bf88.5}\\ 
\hline 
\end{tabular}
\end{center}
\caption{Semantic segmentation results on S3DIS dataset.}
\label{tab:sem_bs}
\end{table}
\vspace{-0.6cm}

Based on the baseline, ASIS associates instance and semantic segmentation, and designs a module to make these two tasks cooperate with each other. Obvious improvements are achieved by ASIS compared with baseline, while our method performes significantly better. Another advantage of our method is that our proposed Self-Prediction head is formulated as a loss function and will be taken off during inference, hence no extra computation burden and space usage are introduced compared with baseline.

\vspace{-0.5cm}
 \vspace{-0.6cm}

\noindent{\textbf{Compare with state-of-the-arts}} We also compare our method with other state-of-the-art methods. Instance segmentation results are shown in Table~\ref{tab:ins_sota}, from which we see that our method achieves state-of-the-art performance. To the best of our knowledge, 3D-BoNet~\cite{DBLP:conf/nips/YangWCHWMT19} is the best published method for instance segmentation in 3D point cloud. Obviously better performance compared with 3D-BoNet is achieved by our method, especially for mean recall. JSNet~\cite{DBLP:conf/aaai/ZhaoT20} achieves excellent performance by designing a feature fusion module based on PointConv~\cite{DBLP:conf/cvpr/WuQL19}. Compared with JSNet, our method (PN++) performs better especially for mCov and mWCov. For semantic segmentation, results are shown in Table~\ref{tab:sem_sota}. Our method achieves comparable results compared with state-of-the-art methods when we only use PointNet++ as backbone. Even better performance on Area 5 is achieved compared with PointCNN, which is an excellent point cloud learning architecture. Moreover, our method is general and can use the most advanced architectures as the backbone to achieve superior performance.
\vspace{-0.5cm}
\begin{table}[!ht]
\begin{center}
\small 
\setlength{\tabcolsep}{6pt}
\begin{tabular}{ccccc}
\hline
\hline
 Method    &  mPrec & mRec  & mCov    & mWCov   \\
\hline
\hline
\multicolumn{5}{c}{Results on Area 5} \\
\hline
  SGPN (PN)~\cite{DBLP:conf/cvpr/WangYHN18}  &  36.0  & 28.7  &  32.7  & 35.5  \\
  3D-BoNet~\cite{DBLP:conf/nips/YangWCHWMT19}&57.5&40.2&-&-\\
  Ours (PN++) & {\bf 60.1} & {\bf 47.2}& {\bf 54.1}  & {\bf 56.3} \\
  
\hline 
\hline
\multicolumn{5}{c}{Results on 6-fold CV} \\
\hline

 SGPN (PN)~\cite{DBLP:conf/cvpr/WangYHN18}& 38.2   & 31.2 & 37.9   & 40.8     \\
 PartNet~\cite{DBLP:conf/cvpr/MoZCYTGS19}&56.4&43.4&-&-\\
 3D-BoNet~\cite{DBLP:conf/nips/YangWCHWMT19}&65.6&47.6&-&-\\
 JSNet~\cite{DBLP:conf/aaai/ZhaoT20}&66.9&53.9&54.1&58.0\\
  Ours (PN++) & {\bf 67.5} & {\bf54.6}  & {\bf 60.4}  & {\bf 63.0} \\
\hline
\end{tabular}
\end{center}
\caption{Instance segmentation results of state-of-the-art methods on S3DIS dataset.}
\label{tab:ins_sota}
\end{table}
\vspace{-1.2cm}

\begin{table}[!hbt]
\small 
\begin{center}
\begin{tabular}{c|ccc|ccc}
\hline
\hline
      Method    & mIoU  & mAcc    & oAcc& mIoU  & mAcc    & oAcc   \\
\hline
\hline 
 &&Area 5&&&6-fold CV \\
\hline
 RSNet~\cite{DBLP:conf/cvpr/HuangWN18} &-&-&-&56.5& 66.5 & -\\
 JSNet~\cite{DBLP:conf/aaai/ZhaoT20} &54.5&61.4&87.7&61.7&71.7&{\bf88.7}\\
  SPGraph~\cite{DBLP:conf/cvpr/LandrieuS18} &  58.0  & 66.5 & 86.5&  62.1& 73.0 &85.5 \\
  PointCNN~\cite{DBLP:conf/nips/LiBSWDC18} &57.3  & 63.9 & 85.9 & {65.4} & {75.6} & {88.1} \\
  PCCN~\cite{DBLP:conf/cvpr/WangSMPU18} &58.3  & {\bf67.0} & -  &-&-&-\\
  PointWeb~\cite{DBLP:conf/cvpr/ZhaoJFJ19} & 60.3& 66.6 &87.0 & {\bf66.7}&{\bf76.2}&87.3 \\
  GACNet~\cite{DBLP:conf/cvpr/WangHHZS19} & {\bf62.9}&- &87.8 &-&-&-\\
 \hline 
 Ours (PN++)&{ 58.8} &{ 65.9}&{\bf 89.2}&{ 64.1} &{ 74.3}&{88.5}\\ 

\hline 
\end{tabular}
\end{center}
\caption{Semantic segmentation results  of state-of-the-art methods on S3DIS dataset.}
\label{tab:sem_sota}
\end{table}
\vspace{-1.3cm}

\subsection{Segmentation Results on ShapeNet}
\begin{figure}[th]
        \begin{center}
        \subfigure[Ins]{\includegraphics[width=0.23\textwidth]{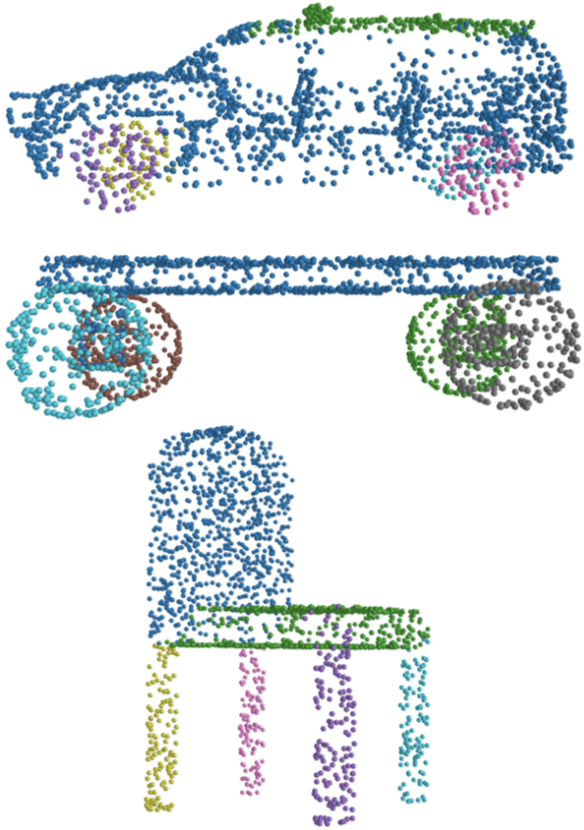}}
        \subfigure[Ins GT]{\includegraphics[width=0.23\textwidth]{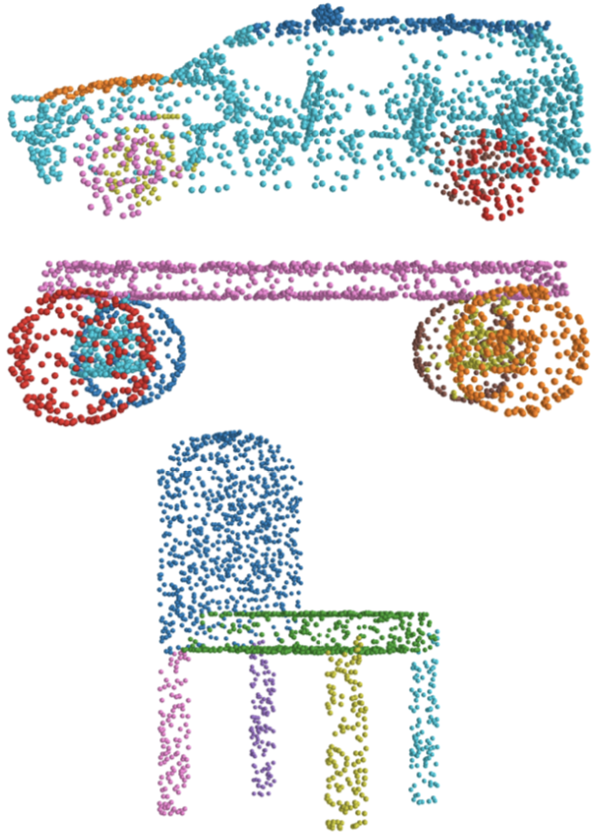}}
        \subfigure[Sem]{\includegraphics[width=0.23\textwidth]{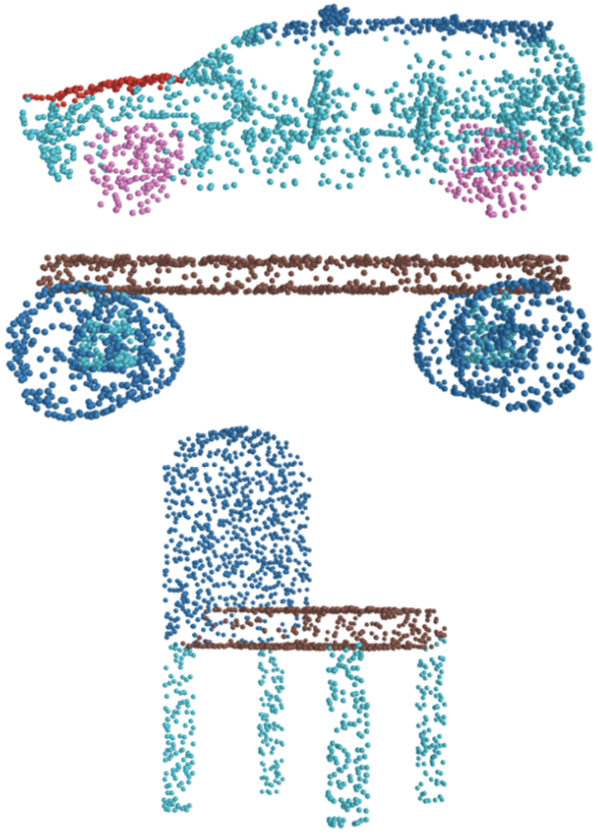}}
         \subfigure[Sem GT]{\includegraphics[width=0.23\textwidth]{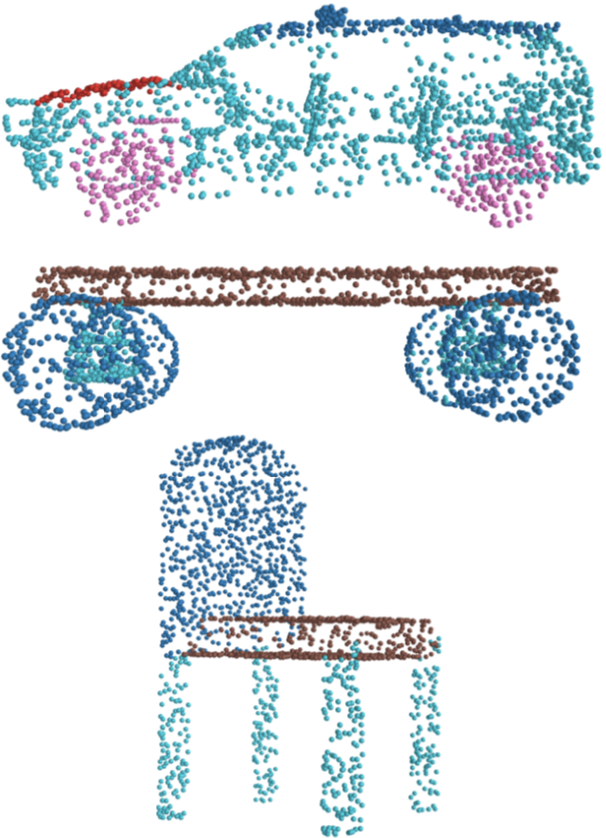}}
         \vspace{-0.2cm}
        \caption{Visualization results of our method on ShapeNet. (a) Instance segmentation results. (b) Instance segmentation ground truth. (c) Semantic segmentation results. (d) Semantic segmentation ground truth.}
        \label{fig:shapenet_vis}
        %\vspace{-0.4cm}
        \end{center}
\end{figure}

We provide qualitative results of part instance segmentation in Figure~\ref{fig:shapenet_vis} following~\cite{DBLP:conf/cvpr/WangYHN18} and~\cite{DBLP:conf/cvpr/WangLSSJ19}. As shown in Figure~\ref{fig:shapenet_vis}, our method successfully segments instances of the same part, such as different legs of the chair. Semantic segmentation results are shown in Table~\ref{tab:sem_sota_sn}, from which we observe that our method achieves significant improvements over baselines. And more improvements compared with ASIS are achieved by our methods. In addition to PointNet and PointNet++, we add a stronger network DGCNN~\cite{dgcnn} in this dataset as our backbone. Obvious performance improvement over baseline also can be observed based on this backbone. 
\vspace{-0.5cm}

\vspace{-1.6cm}
\begin{table}[!ht]
\small
\begin{center}
\begin{tabular}{c|cc}
\hline
 Method       & pIoU &mpIoU\\
\hline
\hline
PointNet (\textit{RePr}) &  83.3 &79.7 \\
PointNet++ (\textit{RePr})&  84.5 &80.5\\
DGCNN (\textit{RePr}) &85.2&82.3 \\
\hline
ASIS (PN) &   84.0 &-\\
ASIS (PN++) &  85.0 &-\\
\hline
Ours (InsSem-SP, PN) &84.5&81.5\\
Ours (InsSem-SP, PN++) &85.8&82.6 \\
Ours (InsSem-SP, DGCNN) &{\bf86.2}&{\bf83.1} \\
\hline
\end{tabular}
\end{center}
\caption{Semantic segmentation results on ShapeNet dataset. \textit{RePr} denotes our reproduced results. All models in the table are trained without normal information.}
\label{tab:sem_sota_sn}
\end{table}

\vspace{-1.4cm}

\subsection{Ablation Study}
In this section, we analyze some important components and hyper-parameters of our methods. All experiments in this section are performed on S3DIS Area 5 using PointNet as backbone. 

\noindent{\textbf{Component Analyses}}. As shown in Section~\ref{sec:results on s3dis}, the effectiveness of our proposed Self-Prediction scheme and joint learning framework has been proved. We further discuss how much our method benefits from bidirectional Self-prediction and class-averaged group dividing way. To this end, two corresponding experiments are conducted: 1) we only perform unidirectional Self-Prediction, and the direction is randomly selected among the two directions, 2) we randomly divide point cloud into groups rather than dividing according to instance labels in Self-Prediction head. Experimental results are reported in Table~\ref{tab:comp_ana}, where mPrec, mRec for instance segmentation and mIoU, mAcc for semantic segmentation are shown. We can draw a conclusion that bidirectional Self-Prediction bring visible improvements compared with unidirectional Self-Prediction and randomly grouping will slightly degrade the performance.
\vspace{-0.5cm}
\begin{table}[!ht]
\begin{center}
\small 
\setlength{\tabcolsep}{6pt}
\begin{tabular}{c|cc|cc}
\hline
\hline
 Method    &  mPrec & mRec  & mIoU    & mAcc   \\
\hline
\hline
 Unidirectional&49.9&40.7&51.0&60.8\\
  Randomly Dividing  &50.5    &42.1   &51.6    &61.3   \\
  Ours (InsSem-SP) & {\bf 51.1} & {\bf 43.6}& {\bf 52.4}  & {\bf 62.4} \\
  
\hline 
\end{tabular}
\end{center}
\caption{Component analyses. Results on S3DIS Area 5 are shown.}
\label{tab:comp_ana}
\end{table}
\vspace{-0.7cm}

\begin{figure}[th]
        \begin{center}
        \subfigure[Analysis of $\beta$]{\label{pa:b} \includegraphics[width=0.31\textwidth]{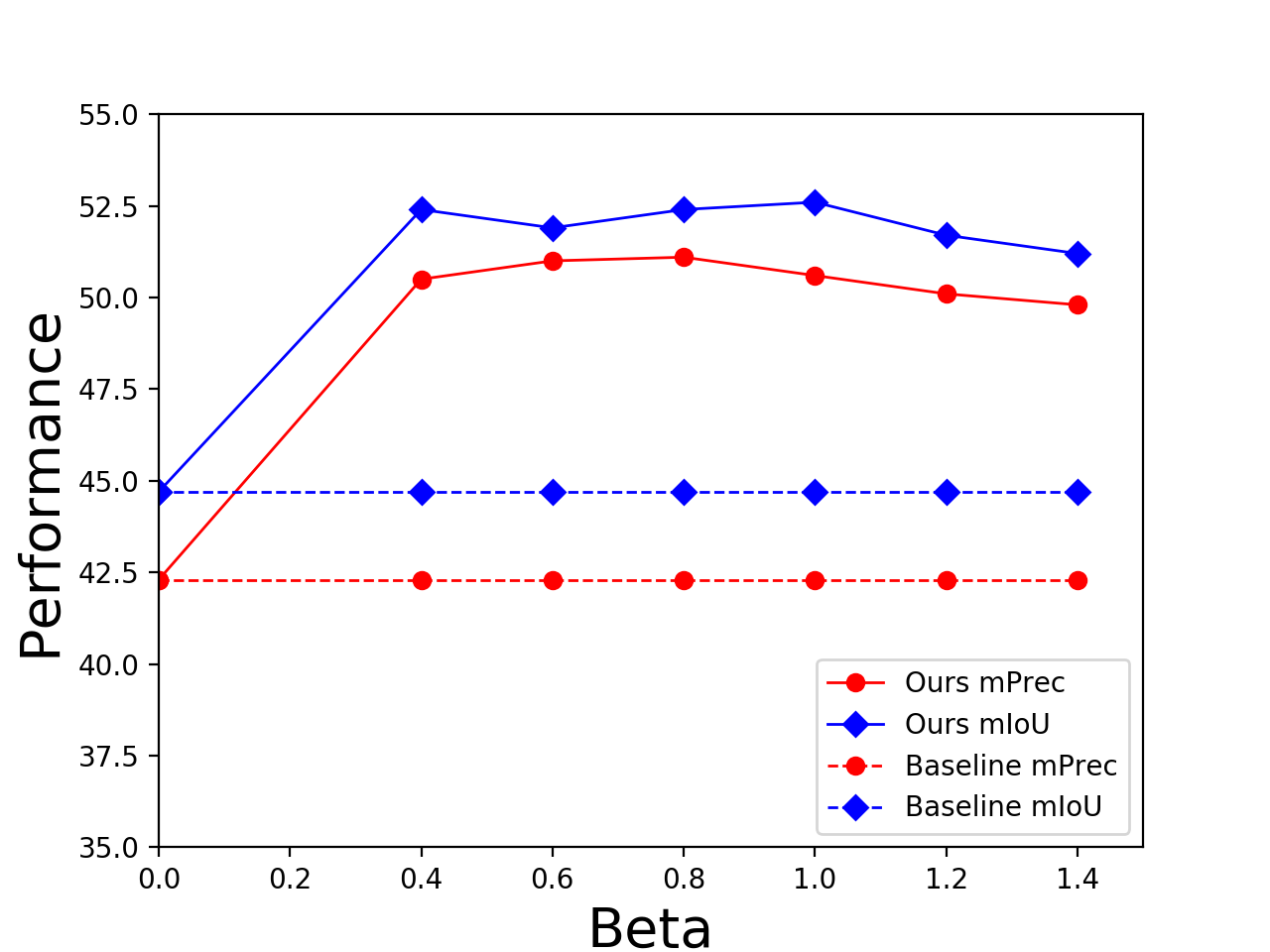}}
        \subfigure[Analysis of $G$]{\label{pa:g} \includegraphics[width=0.31\textwidth]{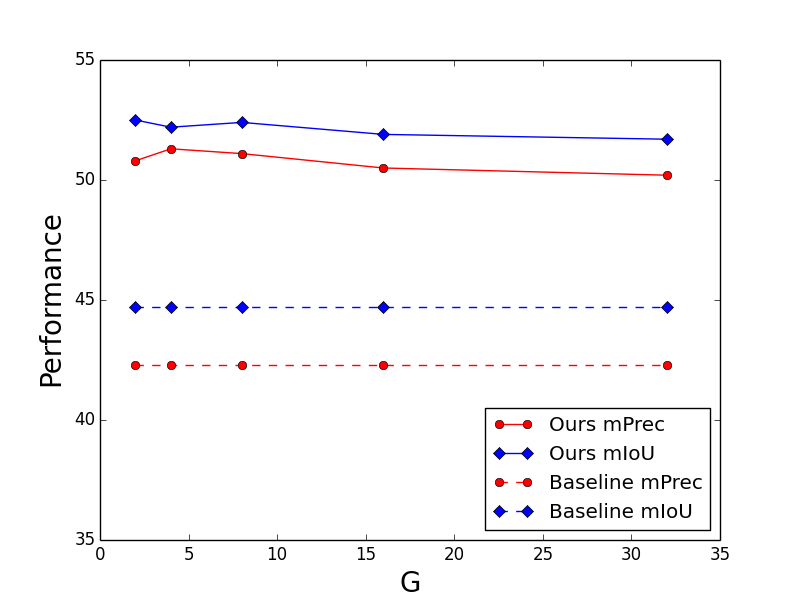}}
        \subfigure[Analysis of $\alpha$]{\label{pa:a} \includegraphics[width=0.31\textwidth]{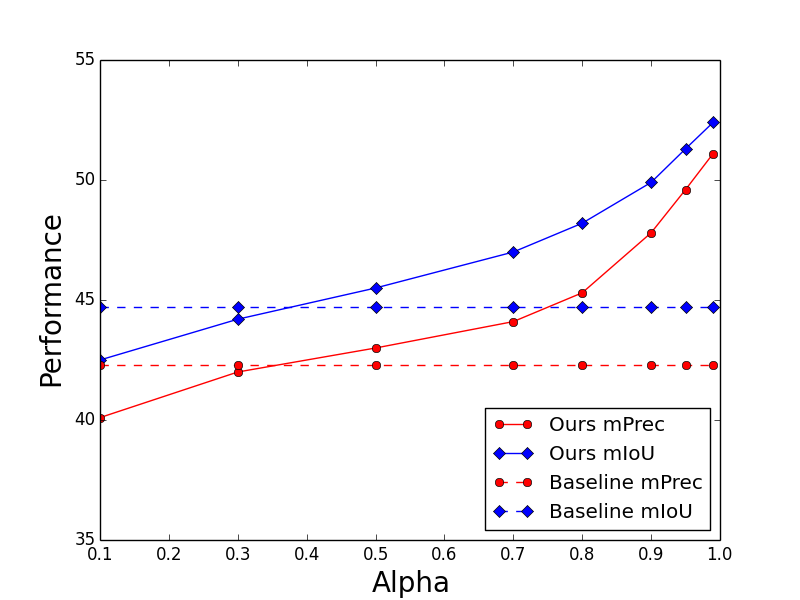}}
        \vspace{-0.2cm}
        \caption{Results of parameter analyses. mPrec for instance segmentation and mIoU for semantic segmentation are shown in figure. Dotted lines represent results of baseline.}
        \label{fig:pa}
        \end{center}
\end{figure}

\noindent{\textbf{Parameter Analyses}}. Three important parameters introduced by our method are analyzed in this section. The first is $\beta$ used to balance the weight of $\mathcal{L}_{sp}$. The analysis results are shown in Figure~\ref{pa:b}, from which we can see that our method is not sensitive to this parameter and works very well in a wide range (0.4-1.4). The second parameter is the number of divided groups $G$ to make a trade off between performance and training speed. We show the results in Figure~\ref{pa:g}, from which we see that the performance is relatively stable and not sensitive to $G$ in a reasonable range. The last parameter is $\alpha$ used to control propagation portion in the process of label propagation. Although we follow the common setting~\cite{DBLP:conf/iclr/LiuLPKYHY19} ($\alpha = 0.99$) in all our experiments, we still conduct experiments to analyze the sensitivity to this parameter of our method. As shown in Figure~\ref{pa:a}, our method outperforms baseline in a large range, i.e., $\alpha > 0.5$.
\vspace{-0.6cm}
\vspace{-1.0cm}
\section{Conclusion}
In this paper, we present a novel learning scheme named Self-Prediction to enforce relation exploring, and a joint framework for associating instance and semantic segmentation of point clouds. Extensive experiments prove that our method can be combined with popular networks significantly improve their performance. By only taking PointNet++ as the backbone, our method achieves state-of-the-art or comparable results. Moreover, our method is a general learning framework and easy to apply to most existing learning networks. 

%\textbf{Acknowledgements.}
%This work was supported by National Science Foundation of China (61976137, U1611461, U19B2035) and STCSM(18DZ1112300).

\clearpage
\bibliographystyle{splncs04}
\bibliography{eccv2020submissionCR}
\end{document}